\documentclass{article}
\usepackage{arxiv}
\usepackage[utf8]{inputenc} 
\usepackage[T1]{fontenc}    
\usepackage{lipsum}   
\usepackage{natbib}
\usepackage{doi}

\usepackage{bm}
\usepackage{subfigure}
\usepackage{subcaption}
\usepackage{caption}
\usepackage{amsfonts}
\usepackage{hyperref}
\usepackage{url}
\usepackage{graphicx}
\usepackage{tabularx}
\usepackage{booktabs}
\usepackage{array}
\usepackage{tcolorbox}
\usepackage{xcolor}
\usepackage{tikz}
\usepackage{booktabs}
\usepackage{mdframed}
\usepackage{xcolor}
\usepackage{pifont}
\usepackage{xspace}
\usepackage{multirow}
\usepackage{wrapfig}
\usepackage{bm}
\usepackage{booktabs}       
\usepackage{amsfonts}       
\usepackage{nicefrac}       
\usepackage{microtype}      
\usepackage{xcolor}    
\usepackage{graphicx}
\usepackage{subfigure} 
\usepackage{amsmath}
\usepackage{amssymb}
\usepackage{makeidx}
\usepackage{multicol}
\usepackage{balance}
\usepackage{enumitem}
\usepackage{array}
\usepackage{comment}
\usepackage{adjustbox}
\usepackage{amsthm}
\usepackage{mwe}
\usepackage{cleveref}    

\begin{document}
\title{Faithful and Interpretable Explanations for Complex Ensemble Time Series Forecasts using Surrogate Models and Forecastability Analysis}
\date{}

\author{
    Yikai Zhao\\
    AWS Supply Chain\\
    \texttt{yikai@amazon.com}
    \and
    Jiekai Ma\\
    AWS Supply Chain\\
    \texttt{jiekal@amazon.com}
}
\maketitle

\begin{abstract}

Modern time series forecasting increasingly relies on complex ensemble models generated by AutoML systems like AutoGluon, delivering superior accuracy but with significant costs to transparency and interpretability. This paper introduces a comprehensive, dual-approach framework that addresses both the explainability and forecastability challenges in complex time series ensembles.
First, we develop a surrogate-based explanation methodology that bridges the accuracy-interpretability gap by training a LightGBM model to faithfully mimic AutoGluon's time series forecasts, enabling stable SHAP-based feature attributions. We rigorously validated this approach through feature injection experiments, demonstrating remarkably high faithfulness between extracted SHAP values and known ground truth effects.
Second, we integrated spectral predictability analysis to quantify each series' inherent forecastability. By comparing each time series' spectral predictability to its pure noise benchmarks, we established an objective mechanism to gauge confidence in forecasts and their explanations. Our empirical evaluation on the M5 dataset found that higher spectral predictability strongly correlates not only with improved forecast accuracy but also with higher fidelity between the surrogate and the original forecasting model. These forecastability metrics serve as effective filtering mechanisms and confidence scores, enabling users to calibrate their trust in both the forecasts and their explanations. We further demonstrated that per-item normalization is essential for generating meaningful SHAP explanations across heterogeneous time series with varying scales.
The resulting framework delivers interpretable, instance-level explanations for state-of-the-art ensemble forecasts, while equipping users with forecastability metrics that serve as reliability indicators for both predictions and their explanations.
\end{abstract}
\section{Introduction}
Time series forecasting is a cornerstone of decision-making across numerous sectors, including demand planning, financial prediction, resource management, and climate modeling. The pursuit of higher predictive accuracy has driven the development of sophisticated machine learning (ML) models, particularly complex ensemble models often generated by Automated Machine Learning (AutoML) frameworks. These ensembles, which combine the strengths of diverse forecasting techniques, have demonstrated significant improvements in predictive performance \citep{Erickson2020AutoGluon}.

However, the very complexity that fuels this enhanced performance often renders these powerful time series ensemble models opaque "black boxes" \citep{Guidotti2018Survey}. In applications where forecasts inform critical decisions, this lack of transparency can erode trust and hinder the adoption of these advanced models \citep{Guidotti2018Survey}. This paper addresses the specific challenge of explaining forecasts generated by such state-of-the-art time series ensemble models.

As a prominent example of such systems, we consider AutoGluon-TimeSeries (AutoGluon-TS). AutoGluon-TS has demonstrated strong empirical performance across 29 benchmark datasets, outperforming a variety of traditional forecasting models in both point and quantile prediction. In many cases, it even exceeds the performance of the best-in-hindsight ensemble of previous methods \citep{Erickson2020AutoGluon}. The high accuracy of AutoGluon-TS, and indeed many cutting-edge time series ensemble systems, is achieved by creating complex combinations of diverse models. These can include statistical methods (like ARIMA, ETS), deep learning models (like DeepAR), and foundational models (like Chronos). This inherent heterogeneity and multi-layered structure make it exceptionally difficult to apply standard Explainable Artificial Intelligence (XAI) techniques directly and reliably across the entire ensemble \citep{Adadi2018Peeking}.

A particular challenge with these complex time series ensembles, as exemplified by AutoGluon-TS, is that their constituent models often require fundamentally different approaches to explanation. This makes it impossible to apply a single explanation method, such as Shapley Additive exPlanations (SHAP), uniformly and meaningfully across all member models. For instance, our preliminary investigations using permutation feature importance directly on an AutoGluon ensemble yielded unstable and inconsistent results in faithfulness tests. This observation aligns with documented limitations of permutation-based methods, which can disrupt the learned relationships between features and the target variable \citep{Hooker2019Unrestricted}.

To overcome these challenges, we propose a surrogate-based explanation methodology. The core idea is to decouple the complex prediction task from the explanation task. We first leverage the power of AutoGluon-TS to train a high-performance forecasting ensemble, treating it as a black box $f_{AG}$. We then train a simpler, inherently more interpretable model – specifically, LightGBM – to act as a surrogate, aiming to accurately mimic the point forecasts generated by the AutoGluon-TS. Once this surrogate model, $f_{LGBM}$, achieves high fidelity in replicating $f_{AG}$'s predictions, we can apply well-established and efficient explanation techniques to it. Additionally, our approach incorporates complementary techniques, such as forecastability analysis, to provide a more robust framework for interpreting complex forecasts and assessing their reliability.

A critical consideration often overlooked in forecast explainability is the inherent forecastability of the time series being analyzed. Explanations derived from models attempting to predict inherently unpredictable or chaotic data may themselves be misleading or unreliable, regardless of the sophistication of the explanation technique. This creates a fundamental challenge: without assessing the intrinsic predictability of the underlying data, users may place unwarranted confidence in explanations of forecasts that are essentially unpredictable. Our framework addresses this challenge by integrating forecastability analysis with model explanations, providing users with critical context about when to trust both forecasts and their explanations. This integration is particularly valuable in business environments where heterogeneous time series with varying levels of predictability must be processed and interpreted efficiently.

The main contributions of this paper are:


\begin{itemize} 
    \item Validation of a surrogate model approach (LightGBM+TreeSHAP) for explaining complex AutoML time series forecasts (e.g., AutoGluon-TS).
    \item Quantitative evaluation of surrogate explanation faithfulness using feature injection, showing high correlation with known ground truth effects.
    \item Integration of spectral predictability and a filter mechanism by comparing it to its white noise benchmarks
    \item Highlighting the necessity and providing a method for per-item normalization enabling comparable SHAP explanations on heterogeneous series.
\end{itemize}

This work aims to offer a practical, validated approach for presenting trustworthy insights into the forecasts of sophisticated ensemble time series models.

\section{Related Work}
This section reviews existing work relevant to explaining time series forecasts, focusing on techniques applicable to surrogate modeling, per-item normalization, and forecastability analysis. 

\subsection{Surrogate Models for XAI}

Surrogate modeling is a practical technique for model-agnostic explainability where an interpretable model (e.g., LightGBM) is trained to mimic a complex black-box system’s input-output behavior \citep{Guidotti2018Survey, Antic2021Principles}. This widely applicable technique continues to be adapted for specific domains, including time series forecasting \citep{Moreno2024sSHAP}, offering valuable simplified global insights into the black box and enhancing comprehension \citep{Antic2021Principles}. Harnessing these benefits critically hinges on ensuring high \textit{fidelity}: the surrogate must accurately replicate the original model's predictions, as explanations lack meaning otherwise \citep{Rudin2019Stop}. Fortunately, fidelity is a quantifiable and verifiable metric, making trustworthy explanation via surrogates a manageable goal rather than an insurmountable barrier. While inherently interpretable models remain ideal \citep{Rudin2019Stop}, surrogates provide an essential bridge when state-of-the-art accuracy necessitates a black-box approach but practical deployment requires understanding its behavior. Consequently, when implemented with rigorous fidelity validation, surrogate modeling offers a robust and highly valuable methodology for practical XAI.


This study adopted LightGBM as the surrogate model, a decision underpinned by two key factors: its consistently demonstrated high performance in forecasting accuracy (e.g., \citep{Ke2017LightGBM}) and its inherent tree-based architecture. This structural characteristic is particularly advantageous as it integrates seamlessly with the efficient TreeSHAP algorithm for the calculation of SHAP values \citep{Lundberg2020TreeSHAP}.

\subsection{Per-Item Normalization for Time Series Explainability}

Per-item normalization is crucial for meaningful time series explainability, particularly with heterogeneous scales common in business forecasting \citep{Molnar2022}. Real-world applications often involve series with vastly different magnitudes, and without normalization, methods like SHAP generate feature attributions biased by absolute scale \citep{Lundberg2017SHAP}.

This scale-dependency fundamentally distorts local explanations across items when applied to unnormalized heterogeneous series. The core issue stems from the SHAP base value ($\phi_0$), which typically represents the average model prediction across the mixed-scale background data. This global $\phi_0$ thus becomes unrepresentative for any specific item whose scale deviates significantly from this overall average. Consequently, the SHAP additivity property ($\hat{y} \approx \phi_0 + \sum \phi_j$) forces the feature contributions ($\sum \phi_j$) to absorb this baseline mismatch. In both low- and high-volume cases, individual SHAP values ($\phi_j$) are disproportionately scaled primarily to compensate for the inappropriate global baseline, rather than accurately reflecting the true, context-specific marginal impact of features relative to that item's own scale. This significantly obscures genuine local insights.

Applying per-item normalization (e.g., $Z$-scoring) addresses this by reframing the task to explain relative deviations from an item-specific baseline. This yields equitable, comparable explanations aligned with business focus on relative impacts \citep{Hoffman2018Metrics}, aiding workflows like exception handling. Conversely, ignoring normalization risks misleading interpretations and suboptimal decisions, undermining trust \citep{Molnar2022}, making per-item normalization a fundamental step for reliable explanations supporting business processes.

\subsection{Time Series Forecastability Analysis}


Real-world business datasets contain heterogeneous time series; some exhibit clear patterns, while others are erratic and resist reliable prediction regardless of model sophistication \citep{Petropoulos2022Review}. Attempting advanced modeling on inherently unpredictable series sets unrealistic expectations for stakeholders and risks misguided decisions based on a false sense of precision. Identifying those time series with a low forecastability score \textit{before} modeling is crucial in large-scale environments like supply chains with numerous item-location combinations, including sporadic ``long-tail'' items, enabling tiered forecasting strategies \citep{Hyndman2011Hierarchical}.

Explanations derived from models attempting to predict inherently chaotic or noisy data may themselves be unreliable \citep{Molnar2022}. Combining forecastability analysis with model explanations yields a more complete picture: understanding not only feature contributions but also whether the underlying data supports reliable prediction. Recent work has focused on quantitative measures to assess time series forecastability prior to modeling:

\textbf{Spectral Predictability:} The spectral predictability score \citep{Goerg2013Spectral} is computed as the entropy of the power spectral density after trend removal. This metric provides a model-agnostic assessment of a time series' intrinsic forecastability.

\textbf{Benchmarking approaches:} Comparing a time series' forecastability metrics against benchmarks derived from noise patterns with equivalent characteristics (length, sparsity) provides a reference point for assessing whether a series contains meaningful signal or is predominantly noise \citep{Kang2017Visualising}. Time series that score below these benchmarks are likely to be inherently unpredictable.

Our work builds on these approaches by integrating spectral predictability analysis with surrogate-based model explanations, creating a comprehensive framework that considers both model behavior and data characteristics when assessing forecast explainability.

\section{Methodology}
The end-to-end process involves training the complex AutoGluon TS model, engineering relevant and explainable features, training a LightGBM surrogate to mimic the normalized AutoGluon predictions, applying TreeSHAP to the surrogate, evaluate faithfulness of the SHAP values derived from the surrogate model, optionally caliberate the SHAP values to be additive to the AutoGluon's raw forecasts\footnote{Details on the calibration method are provided in Appendix.} and finally provide explanation based on SHAP values. This process could be augmented by forecastability as a filter or confidence score mechanism for the forecasts and explanations.

\subsection{Feature Engineering for Explainability}

We engineered a distinct feature set ($X_{eng}$) as interpretable input for the LightGBM surrogate model. This set differs from features typically used by complex forecasters like AutoGluon by deliberately limiting extensive lag features (to avoid diluting SHAP explanations) while augmenting with other signals. This comprehensive set comprised: time-based features (week, month, day of week, etc.), key target lags, rolling/expanding window statistics (mean, std dev, skewness, etc.), percentage changes, signal decomposition features (trend, seasonality), and history/age features. These features were chosen to provide interpretable drivers for the explanation model.

\subsection{Per-Item Normalization Strategy}
Given the wide variation in demand scales in the M5 dataset, applying a consistent normalization scheme per-item is crucial before training the surrogate and calculating SHAP values.

We employ per-item standardization (z-score normalization). For each item $i$, we calculate the mean $\mu_i$ and standard deviation $\sigma_i$ of its historical demand $y_i$. The target variable for the surrogate model becomes the normalized AutoGluon forecast: $z_{AG,k}=(\hat{y}_{AG,k}-\mu_i)/\sigma_i$. The LightGBM model $f_{LGBM}$ is trained to predict $z_{AG,k}$. SHAP values $\phi_k$ are then computed to explain $f_{LGBM}(x_{eng,k}) \approx z_{AG,k}$. For final interpretation in the original demand units, the SHAP values can be denormalized by multiplying by $\sigma_i$.



\subsection{Faithfulness Evaluation}
\label{sec:faithfulness_eval} 

Faithfulness ensures explanations accurately reflect a model's reasoning or true feature influence, a critical check beyond predictive fidelity \citep{Molnar2022, Yeh2019Faithfulness, Rudin2019Stop}. Evaluating faithfulness rigorously is challenging due to the typical absence of ground truth explanations for real-world models \citep{Doshi-Velez2017}. A key strategy to overcome this, recognized in recent evaluation benchmarks \citep{Li2023M4}, is the use of synthetic or known ground truths; this principle has been practically applied, for instance, to evaluate anomaly explanations against generated ground truths \citep{Angelov2023Evaluating}. Our work employs a specific form of this strategy: a feature injection experiment, conceptually illustrated in the Appendix (Table~\ref{tab:feature_injection_example}). This approach also serves as a practical debugging test and sanity check for explanation reliability \citep{Slack2022Debugging, Adebayo2018SanityChecks}. The method involves introducing a synthetic feature with a known, predefined impact on the target variable, retraining the surrogate model on this modified data, and extracting the explanation (e.g., SHAP values, $\phi_{\textit{injected}}$) for the synthetic feature. Faithfulness is then quantified by comparing the extracted explanation $\phi_{\textit{injected}}$ against the known ground truth effect; high correlation indicates the method accurately captures the injected influence \citep{Molnar2022}.

This technique provides a valuable quantitative faithfulness measure for at least one feature's effect and serves as a crucial debugging tool \citep{Slack2022Debugging, Doshi-Velez2017}. However, its primary limitation is that it mainly validates the explanation for the injected feature's main effect, potentially not fully capturing faithfulness regarding complex interactions among the original features \citep{Slack2022Debugging, Molnar2022, Yeh2019Faithfulness}. Despite this, positive results build confidence and provide an important sanity check, aligning with established principles for evaluating XAI against known or synthetic truths \citep{Li2023M4, Angelov2023Evaluating}.

\subsection{Forecastability Analysis Implementation}

To provide context for forecast and explanation reliability, we quantified the intrinsic predictability of each time series using the Spectral Predictability (SP) approach, assessing both its average level and its stability over time.

\textbf{1. Preprocessing:} Each series was preprocessed by removing leading zeros, yielding processed series $y'_{i}$ with effective length $L_i$. The effective length $L_i$ will be at least 4 times of prediction length based for the forecasting purpose. 

\textbf{2. Average SP \& Benchmark Comparison:} The overall SP score, $SP_i$, was computed for each processed series $y'_{i}$. To account for SP's length dependency, this score was compared against a length-specific pure noise benchmark, $SP_{noise}(L_i)$. Series with $SP_i$ near or below this benchmark were flagged, suggesting predictability close to random noise.


\section{Results}
This section presents and discusses the results from our experiments to evaluate the proposed framework. The experiments were developed using the M5 Forecasting Competition dataset\footnote{Dataset available at: \url{https://www.kaggle.com/competitions/m5-forecasting-accuracy/data}}, a real-world hierarchical collection featuring daily sales data from Walmart stores. This dataset spans 1,941 days and encompasses multiple organizational levels, including 10 stores, 7 departments, and 3 categories, tracking a total of 3,049 unique items. To determine the most appropriate forecasting level for our experiments, we conducted the spectral predictability analysis across different hierarchical levels, summarized in the Appendix (Table~\ref{tab:m5_predictability}). This analysis indicated that the store-department level (70 time series) provided a strong balance of aggregation and inherent forecastability (approximately 94\% of series deemed forecastable), leading us to select this granularity.
Our study focuses on this subset of 70 store-department daily demand time series. An initial exploration revealed significant heterogeneity in demand scales across different store-department combinations, with mean daily demand varying considerably. This observation strongly motivated the need for a per-item (per-series) normalization strategy before modeling and explanation.
For our experimental setup, we utilized the first 1,913 days of data as the training set and performed forecasting for the subsequent 28-day horizon. The actual data from days 1,914 to 1,941 (inclusive) served as the test set for forecast accuracy validation and explanation analysis.

\subsection{Surrogate Model Fidelity}

\begin{figure}[h]
\centering
\includegraphics[width=\linewidth]{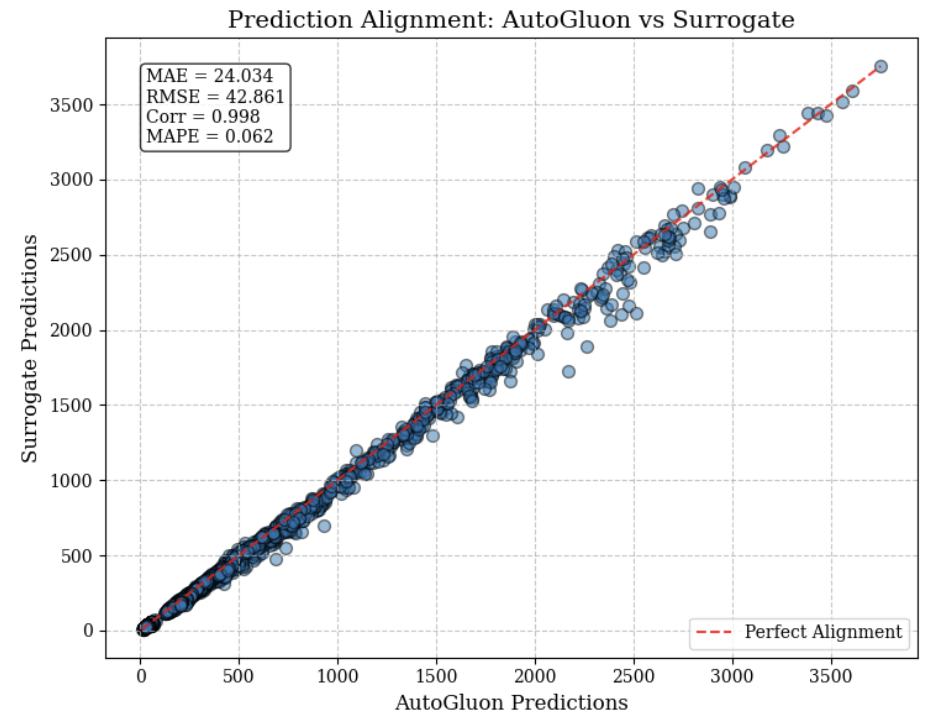} 
\caption{Prediction Alignment between AutoGluon Model and Surrogate Model}
\label{fig:fidelity_scatter}
\end{figure}

We first assessed the fidelity of a LightGBM surrogate in approximating the normalized point forecasts produced by the high-performing AutoGluon ensemble. Fidelity was quantified by comparing predictions from both models on the test set using per-series metrics, including mean absolute error (MAE), mean absolute percentage error (MAPE), and root mean squared error (RMSE). Figure~\ref{fig:fidelity_scatter} illustrates a good alignment, with predicted values tightly clustered around the identity line. This high fidelity enables the use of the complex AutoGluon model for accurate forecasting, while facilitating post hoc interpretability via TreeSHAP applied to the computationally efficient LightGBM surrogate. This fidelity check is a first sanity check for such XAI system.

\subsection{Surrogate Model Faithfulness}

\begin{figure}[h]
\centering
\includegraphics[width=\linewidth]{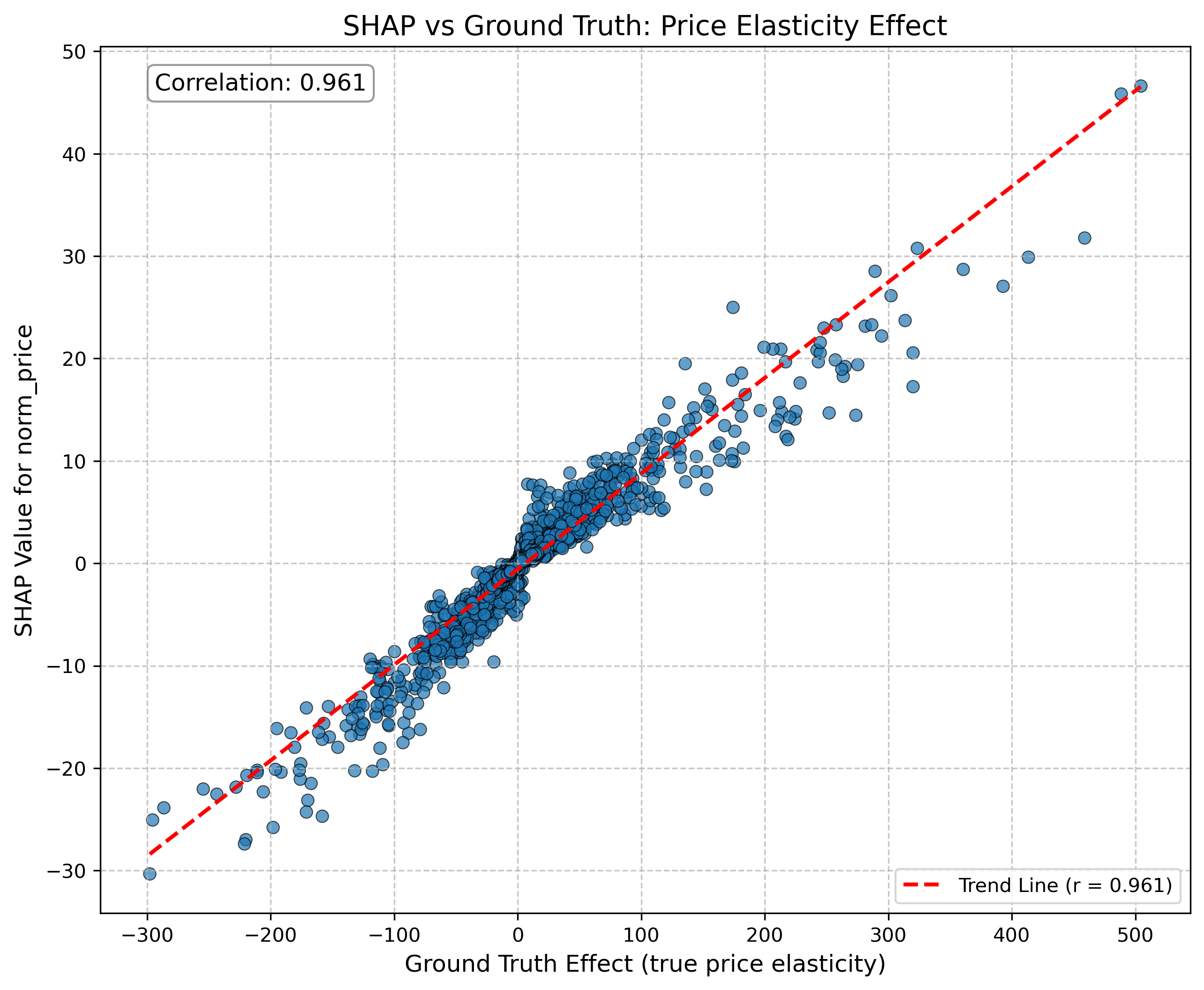} 
\caption{SHAP Faithfullness Validation by Price Effects}
\label{fig:faithfulness_plot}
\end{figure}

To further validate the reliability of the generated explanations, we evaluated explanation faithfulness through a feature injection experiment. This evaluation compared the denormalized SHAP values derived from the surrogate model for an injected feature ($\phi_{price}$) against its known ground truth effect. The results demonstrated high faithfulness, with a Pearson correlation coefficient of 0.961 between the extracted SHAP values and the ground truth impacts. Figure~\ref{fig:faithfulness_plot} visually reinforces this result, showing a strong positive linear relationship. This high correlation validates faithfulness by confirming that the SHAP values correctly track the direction and relative magnitude of the known influence, even if the absolute scales on the plot differ due to the specific ground truth construction artifact. This evidence substantiates that the surrogate + SHAP framework can accurately recover the influence of individual features when the true effects are known, thereby reinforcing the trustworthiness of the explanations regarding feature effects. Nonetheless, as noted in prior discussions on the limitations of faithfulness evaluation, the reliable attribution of interaction effects remains an unresolved challenge. As such, generalizing the observed faithfulness from the injected feature's main effect to all original features warrants caution.

\subsection{Necessity of Per-item Normalization}

Our experiments empirically validated the critical necessity of per-item normalization discussed previously. For example, consider the items whose average demand varies by orders of magnitude, as shown by example items in the Appendix (Table~\ref{tab:norm_example_items}).
As visually demonstrated for a low-volume item in the Appendix (Figure~\ref{fig:norm_shap_comparison}), applying SHAP without normalization results in distorted explanations. The large global base value forces compensatory SHAP values that obscure the true local feature impacts relative to the item's specific scale. In contrast, the results confirmed that per-item normalization yields explanations reflecting relative feature importance against an appropriate item-specific baseline, enabling meaningful interpretation and comparison across series with differing scales.

\subsection{Forecastability Analysis} 

Beyond evaluating fidelity and faithfulness, we further analyzed relationship between forecastability and forecast accuracy and surrogate model fidelity. These results underscore the critical role of data properties in both generating reliable forecasts and interpreting model explanations.



Our analysis revealed strong positive correlations between average spectral predictability and both AutoGluon-TS forecast accuracy (Figure~\ref{fig:accuracy_vs_sp}) and surrogate model fidelity (Figure~\ref{fig:fidelity_vs_sp} in the Appendix). This demonstrates that time series with more regular, predictable patterns not only yield more accurate forecasts but also enable more faithful mimicry by the surrogate model. The relationship becomes particularly evident at the lower end of the spectrum—both accuracy and fidelity deteriorate significantly for series with spectral predictability approaching or below the pure noise baseline. These findings highlight the practical value of spectral predictability as a diagnostic metric, offering a reliable mechanism to filter potentially problematic series or flag when explanations are generated for forecasts that may be inherently compromised by the unpredictable nature of the underlying data.


\begin{figure}[h]
\centering
\includegraphics[width=\linewidth]{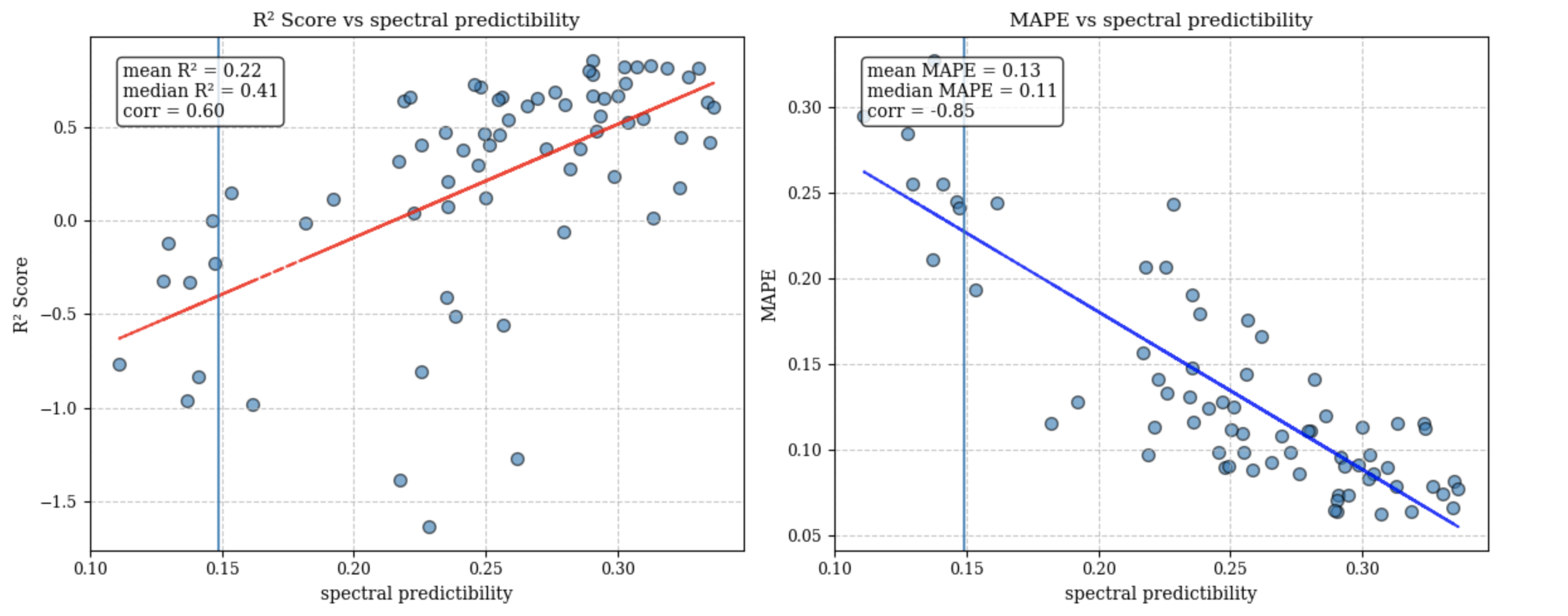} 
\caption{AutoGluon Forecast Accuracy vs. Average Spectral Predictability. Left: using $R^2$ metric. Right: using MAPE metric.}
\parbox{\columnwidth}{\small \textit{Note:} The vertical line indicates the average spectral predictability score for pure noise, serving as a benchmark for randomness.}
\label{fig:accuracy_vs_sp}
\end{figure}

\subsubsection{Case Examples: High vs. Low Forecastability} 

To illustrate the practical implications of forecastability, we examine two contrasting case studies, comparing the alignment of AutoGluon forecasts, surrogate model forecasts, and actual demand.


Illustrative case studies, detailed in the Appendix~(Section~\ref{app:case_studies}), contrast a high-forecastability series (\texttt{HOUSEHOLD\_1} at \texttt{CA\_3}, Appendix Figure~\ref{fig:High SP Alignment}) with a low-forecastability one (\texttt{HOBBIES\_2} at \texttt{TX\_1}, Appendix Figure~\ref{fig:Low SP ALignment}). The high-forecastability example demonstrates close tracking between actual demand, AutoGluon-TS predictions, and surrogate forecasts, indicating high forecast accuracy and surrogate fidelity. In contrast, the low-forecastability example exhibits poorer accuracy for both models and noticeable divergence between the AutoGluon-TS and surrogate predictions, signifying lower fidelity. These cases visually confirm that higher intrinsic data predictability (context in Appendix Figures~\ref{fig:High SP benchmark} and \ref{fig:Low SP benchmark}) supports both better forecast performance and more faithful surrogate model replication.

These examples visually reinforce the link between a time series' intrinsic forecastability and the performance achievable by both complex forecasting models and the surrogate models intended to explain them. High forecastability correlates positively with both forecast accuracy and surrogate fidelity, whereas low forecastability often leads to reduced performance on both fronts, highlighting the importance of considering data characteristics when interpreting model outputs and their explanations.

These empirical findings strongly advocate for integrating forecastability analysis directly into the explainability workflow. Quantitative forecastability metrics offer significant practical value by serving as an essential filter or confidence score when interpreting explanation results \citep{Kang2017Visualising, Petropoulos2022Review}. By assessing metrics against predefined benchmarks (e.g., derived from noise), we can identify series where inherent data characteristics likely undermine model reliability. Low forecastability signals potential issues: the underlying forecast being explained may be inaccurate, the surrogate model might struggle to achieve high fidelity, and the feature relationships captured by the explanation could be unstable or noise-driven. Consequently, explanations associated with low-forecastability scores warrant lower confidence and can be appropriately flagged or filtered in practical applications. Explicitly acknowledging inherent data limitations via forecastability analysis thus provides a crucial layer of understanding. This helps manage user expectations regarding prediction certainty and prevents over-reliance on model outputs or their explanations when the underlying data simply does not support high confidence \citep{Molnar2022}. Ultimately, this combined framework of explainability and forecastability enables a more nuanced, trustworthy, and responsible interpretation of advanced forecasting systems.

\section{Conclusion}

This paper addressed the critical challenge of explaining complex ensemble time series forecasting models, specifically those generated by AutoML systems like AutoGluon, where direct explanation methods often prove unstable or infeasible. We proposed and validated a surrogate-based methodology, training a LightGBM model to mimic AutoGluon’s point forecasts with high fidelity. We generated stable local feature attributions, whose faithfulness was confirmed via a rigorous feature injection experiment. We also demonstrated the crucial role of per-item normalization in enabling meaningful interpretations across heterogeneous time series. This work highlighted the significant benefit achieved by integrating these surrogate-based explanations with forecastability analysis (e.g., using spectral predictability). This combination allows users to calibrate their trust in an explanation based on the inherent predictability, which we found directly correlates with both improved forecast accuracy and higher surrogate model fidelity, or lack thereof, in the underlying time series. This synergy bridges the gap between understanding model behavior and acknowledging data limitations, preventing over-reliance on potentially misleading explanations when data characteristics inherently limit predictive certainty. Such integrated approaches, balancing model interpretability with an understanding of data forecastability, are vital for the transparent and responsible deployment of advanced forecasting systems.

\newpage 
\appendix

\section{Predictability Analysis Results} 
\label{app:predictability}


\begin{table}[htbp]
\centering
\caption{M5 Dataset Predictability Analysis by Hierarchical Level}
\label{tab:m5_predictability}
\small 
\begin{tabular}{@{}lrrr@{}}
\toprule
Level            & \# Time Series & \# Forecastable Series & Forecastable (\%) \\
\midrule
Store            & 10             & 10                     & 100.0\%           \\
Department       & 7              & 7                      & 100.0\%           \\
Store-Department & 70             & 66                     & 94.3\%            \\
Product          & 3049           & 496                    & 16.3\%            \\
\bottomrule
\end{tabular}
\vspace{2mm}
\parbox{\columnwidth}{\small \textit{Note:} A time series was deemed `forecastable' if its mean spectral predictability score was higher than its baseline score. The implementation of the spectral predictability analysis is detailed in the main text.} 
\end{table}

\section{Per-Item Normalization Illustration}
\label{app:normalization}


\begin{table}[h]
\centering
\caption{Example Items with Different Demand Scales}
\label{tab:norm_example_items}
\small 
\begin{tabular}{@{}lcc@{}} 
\toprule
Item ID & Mean Weekly Demand & Std Dev Weekly Demand \\ 
\midrule
Item 0 (High Vol) & 1179.86 & 315.96 \\
Item 1 (Med Vol)  & 116.80  & 26.83  \\
Item 2 (Low Vol)  & 11.66   & 2.84   \\
\bottomrule
\end{tabular}
\end{table}

\begin{figure}[h]
\centering
\includegraphics[width=\linewidth]{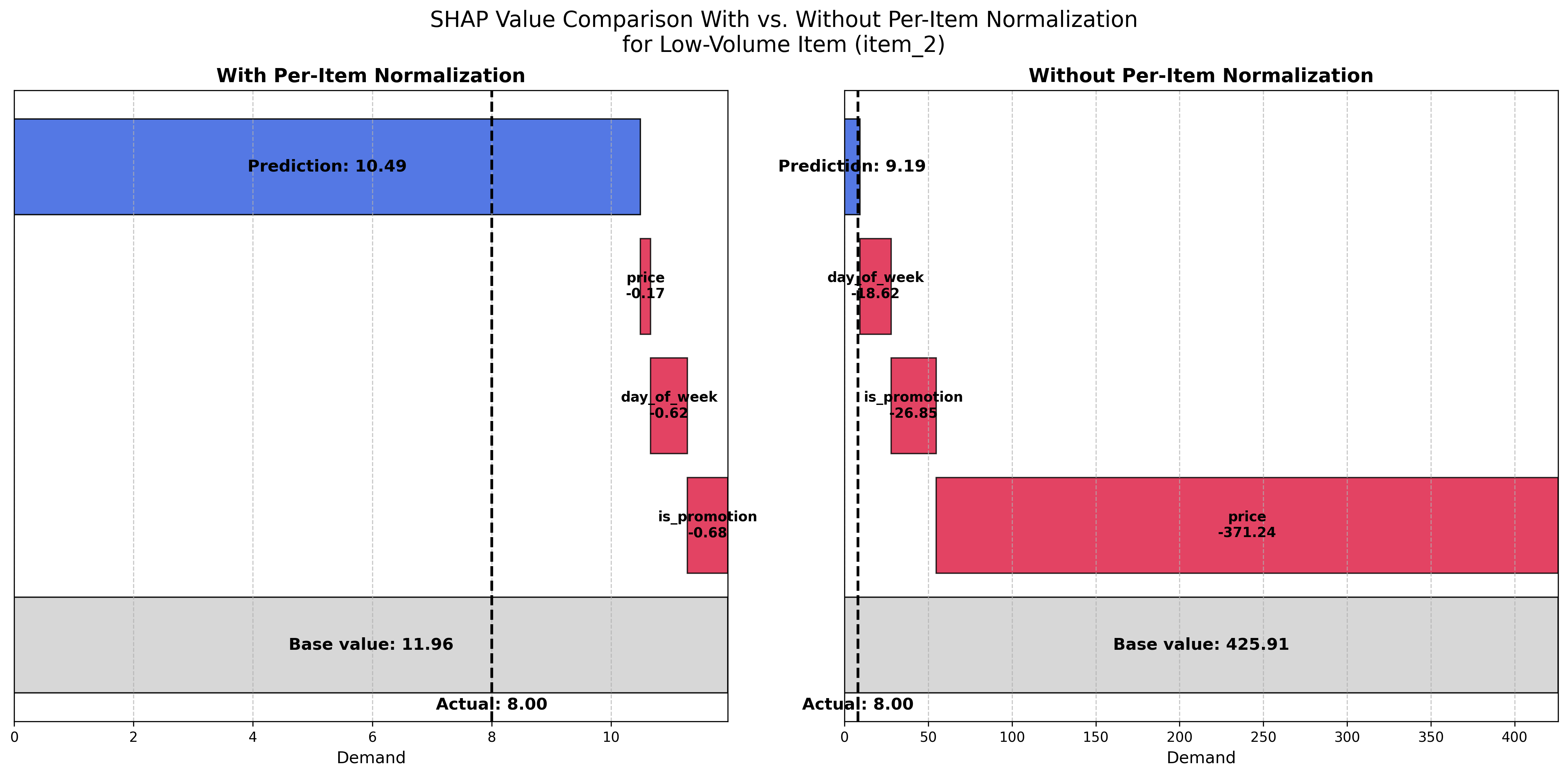}
\caption{Conceptual Illustration of SHAP Values With and Without Per-Item Normalization for Items of Different Scales.}
\label{fig:norm_shap_comparison}
\end{figure}

\section{Correlation Between Spectral Predictability and Fidelity}
This section places the correlation plot regarding Spectral Predictability and fidelity. The plot has been referenced in the main body of the paper.

\begin{figure}[h]
\centering
\includegraphics[width=\linewidth]{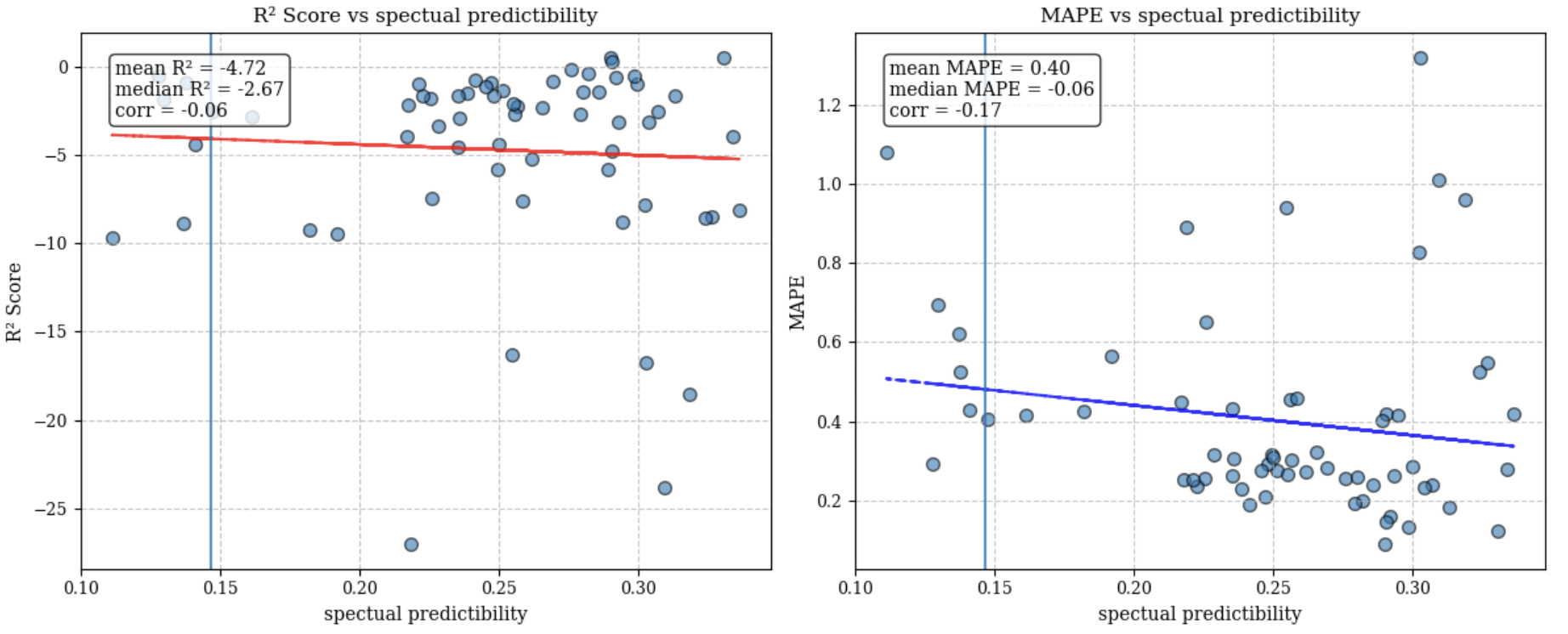} \caption{Surrogate Model Fidelity vs. Average Spectral Predictability. Left: using $R^2$ metric. Right: using MAPE metric.}
\label{fig:fidelity_vs_sp}
\end{figure}

\section{Example of High vs Low Forecastability} 
\label{app:case_studies}

This section provides the SHAP Explanations of two department-stores' forecasting as the illustrution. The HOUSEHOLD\_1 of CA\_3 store is highly forecastable (high spectral predictibility), while HOBBIES\_2 at TX\_1 store has low spectral predictibility. 

First, consider \texttt{HOUSEHOLD\_1} at \texttt{CA\_3}, identified as a series with relatively high forecastability, as shown in Figure~\ref{fig:High SP benchmark} in the Appendix. Figure~\ref{fig:High SP Alignment} in the Appendix displays the forecast results. The plot clearly shows strong, regular patterns in the actual demand (green line). This plot indicates high forecast accuracy for both the AutoGluon-TS ensemble in this high-forecastability scenario. Furthermore, the surrogate's predictions align tightly with AutoGluon-TS's predictions, demonstrating high fidelity.

In contrast, Figure~\ref{fig:Low SP ALignment} in the Appendix presents results for \texttt{HOBBIES\_2} at \texttt{TX\_1}, a series characterized by a relatively low forecastability, as shown in Figure~\ref{fig:Low SP benchmark} in the Appendix. The actual demand here is much more erratic and lacks the clear seasonality seen previously. Consequently, both the AutoGluon and surrogate forecasts exhibit lower accuracy, struggling to capture the sharp peaks and troughs in the actual demand, although they follow the general level. Critically, there are noticeable divergences between the AutoGluon and surrogate model forecasts at several points across the horizon. This visual gap signifies reduced surrogate fidelity compared to the high-forecastability example, confirming that it is more challenging for the surrogate to mimic the  AutoGluon-TS's behavior when the underlying series is inherently less predictable.


\begin{figure}[h]
\centering
\includegraphics[width=\linewidth]{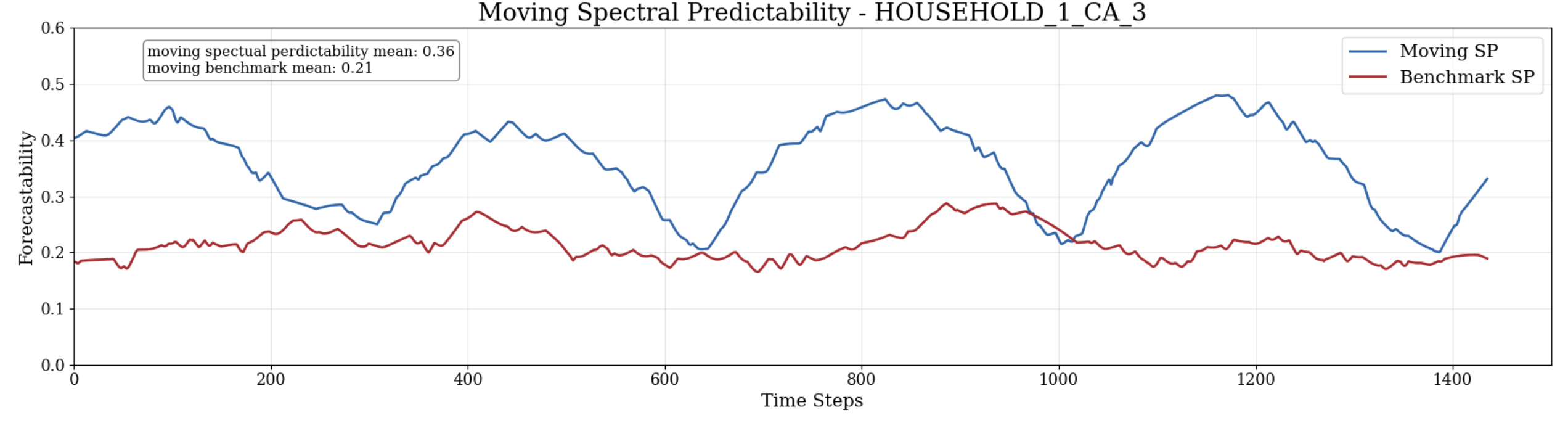} 
\caption{spectral Predictibility vs. Benchmark (random noise) for HOUSEHOLD\_1 at CA\_3 Store }
\label{fig:High SP benchmark}
\end{figure}

\begin{figure}[h]
\centering
\includegraphics[width=\linewidth]{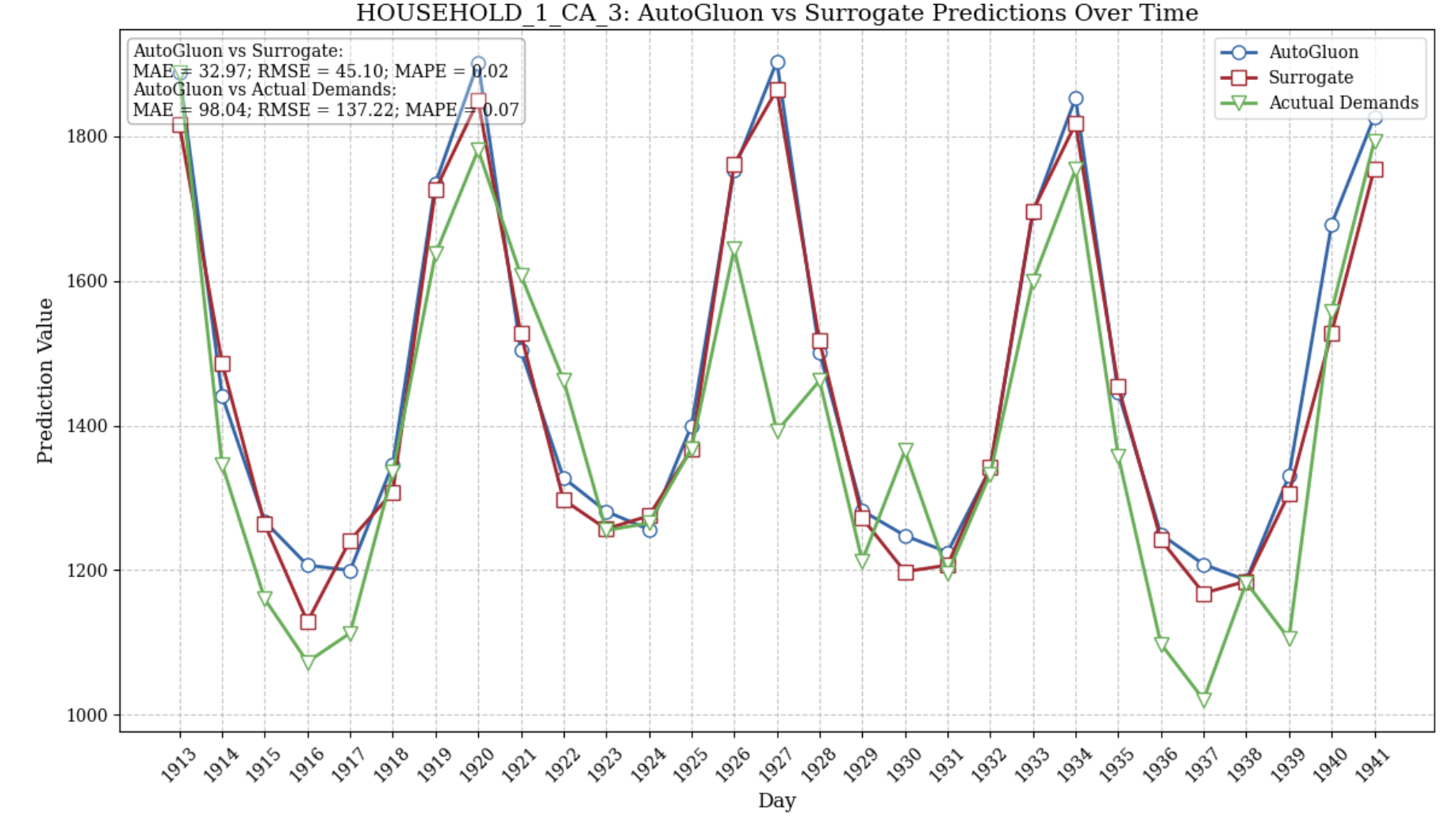} 
\caption{Comparison of Surrogate and AutoGluon Forecasts for HOUSEHOLD\_1 at CA\_3 Store}
\label{fig:High SP Alignment}
\end{figure}

\begin{figure}[h]
\centering
\includegraphics[width=\linewidth]{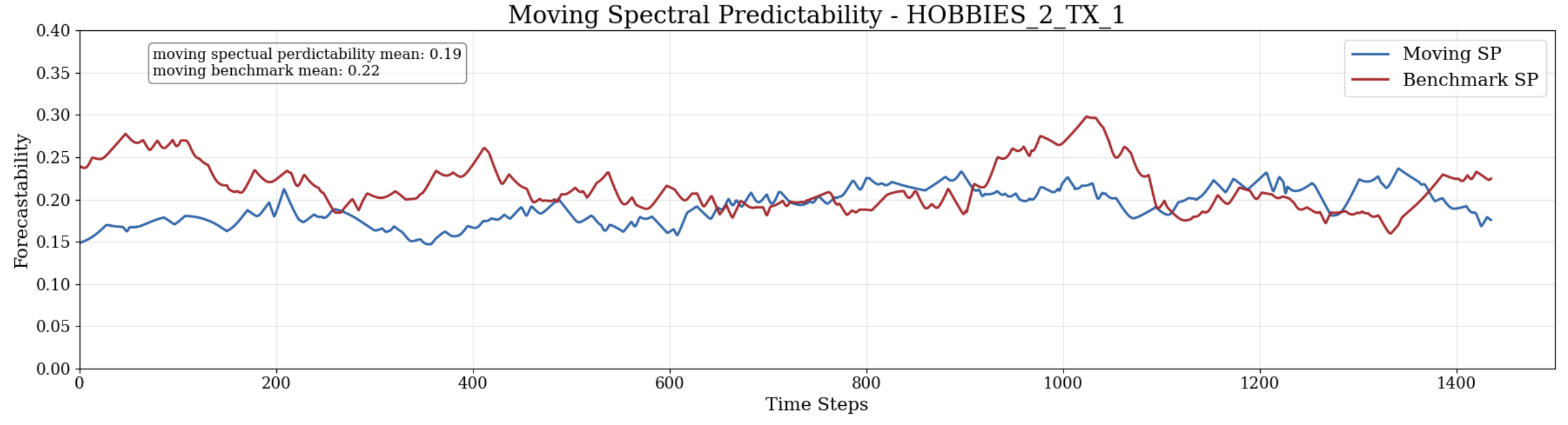} 
\caption{spectral Predictibility vs. Benchmark (random noise) for HOBBIES\_2 at TX\_1 Store }
\label{fig:Low SP benchmark}
\end{figure}

\begin{figure}[h]
\centering
\includegraphics[width=\linewidth]{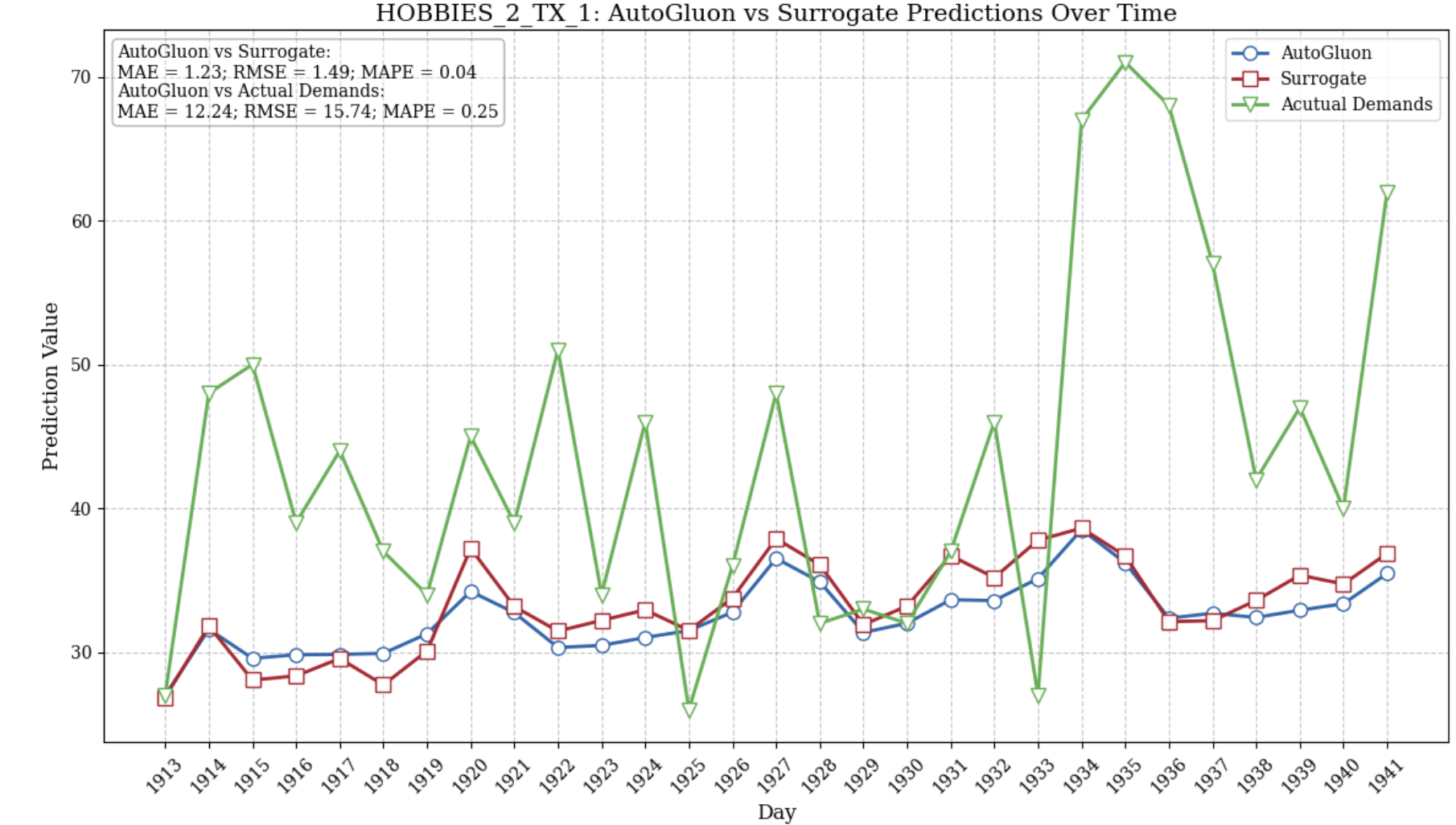} 
\caption{Surrogate vs. Auto Gluon Forecasting for HOBBIES\_2 at TX\_1 Store forecasting}
\label{fig:Low SP ALignment}
\end{figure}


\section{Example of Feature Injection Faithfulness Test} 
This section provides a conceptual illustration of the feature injection experiment used to evaluate explanation faithfulness, as discussed in Section~\ref{sec:faithfulness_eval}. This method allows for quantitative assessment by creating a known ground truth effect for at least one feature, helping to validate if explanation techniques accurately capture feature influence.

\label{app:feature_injection}

\begin{table}[htbp]
\centering
\caption{Conceptual Example of Feature Injection: Simulating Price Effect on Demand}
\label{tab:feature_injection_example}
\footnotesize
\setlength{\tabcolsep}{2.5pt}

\begin{tabular*}{\columnwidth}{@{\extracolsep{\fill}} llrrrr @{}}
\toprule
\textbf{ID} & \textbf{Date} & \shortstack{\textbf{Price}\\\textbf{(\$)}} &
\shortstack{\textbf{Price Effect}\\\textbf{(On Demand)}} &
\shortstack{\textbf{Old}\\\textbf{Demand}} &
\shortstack{\textbf{New}\\\textbf{Demand}} \\
\midrule
\texttt{ITEM\_A} & \texttt{2025-05-01} & 10.00 &   0 & 100 & 100 \\
\texttt{ITEM\_A} & \texttt{2025-05-02} & 11.00 & -10 & 102 &  92 \\
\midrule
\texttt{ITEM\_B} & \texttt{2025-05-01} & 20.00 &   0 &  50 &  50 \\
\texttt{ITEM\_B} & \texttt{2025-05-02} & 21.00 &  -5 &  51 &  46 \\
\texttt{ITEM\_B} & \texttt{2025-05-03} & 19.00 &  +5 &  49 &  54 \\
\bottomrule
\end{tabular*}

\vspace{2mm}
\end{table}


\section{Calibration}
To potentially improve alignment between the explanation and the original model's prediction \texttt{$\hat{y}_{AG}$}, especially where the surrogate \texttt{$f_{LGBM}$} deviates significantly, a post-processing calibration step can be considered. Note this is an optional step and lacks any theoretical support. It is meant as a patch for a system requirements that must bridge the gap between the surrogate model's forecast and that of the AutoGluon TS. 
Acknowledging the SHAP additivity property (\texttt{$f_{LGBM}(x) = \phi_0 + \sum_j \phi_j$}, where \texttt{$\phi_0$} is the base value from the surrogate explanation and \texttt{$\phi_j$} are the surrogate's SHAP values \citep{Lundberg2017SHAP}), a more appropriate calibration aims to rescale the feature contributions (\texttt{$\phi_j$}) so their sum matches the deviation of the original prediction from the base value (\texttt{$\hat{y}_{AG} - \phi_0$}).

This can be achieved heuristically by calculating a scaling factor $s$:
$$ s = \frac{\hat{y}_{AG} - \phi_0}{f_{LGBM}(x) - \phi_0} = \frac{\hat{y}_{AG} - \phi_0}{\sum_j \phi_j} $$
assuming the denominator ($\sum_j \phi_j$) is non-zero. The calibrated SHAP values are then $ \phi_{j,calib} = s \times \phi_j $. This factor $s$ naturally handles potential sign inversions when $\phi_0$ lies between $f_{LGBM}(x)$ and $\hat{y}_{AG}$.

Handling edge cases, such as $\hat{y}_{AG} = 0$ while $f_{LGBM}(x) \neq 0$, requires careful consideration:
\begin{itemize}
    \item \textbf{If surrogate predicts base value ($\boldsymbol{f_{LGBM}(x) = \phi_0}$):}
        \begin{itemize}
            \item If the target model also predicts the base value ($\boldsymbol{\hat{y}_{AG} = \phi_0}$), both models agree on zero deviation from the baseline. No calibration is needed; the original (zero-sum) surrogate SHAP values $\phi_j$ can be used.
            \item If the target model predicts differently ($\boldsymbol{\hat{y}_{AG} \neq \phi_0}$), calibration is mathematically impossible (division by zero: $\sum_j \phi_j = 0$). This indicates a significant local mismatch where the surrogate fails to capture the target's deviation. Action: Issue a warning about the fidelity failure and either skip displaying calibrated values or show the original $\phi_j$ with a strong caveat.
        \end{itemize}
    \item \textbf{If target model predicts zero ($\boldsymbol{\hat{y}_{AG} = 0}$):}
        \begin{itemize}
            \item Calibration can generally proceed using the formula $s = -\phi_0 / (f_{LGBM}(x) - \phi_0)$, as long as the surrogate prediction is not exactly the base value (i.e., $f_{LGBM}(x) \neq \phi_0$).
            \item However, consider the practical context: for zero forecasts often driven by specific business rules (e.g., stock-outs, discontinued items), providing a rule-based explanation might be more insightful than displaying potentially complex calibrated SHAP values that sum to $-\phi_0$.
        \end{itemize}
\end{itemize}

We acknowledge that any such calibration remains a heuristic adjustment applied post-hoc \citep{Molnar2022}. It lacks strong theoretical grounding within the SHAP framework, as it modifies explanation values based on outcomes rather than directly reflecting the (surrogate) model's internal logic that generated the original $\phi_j$. It should be applied cautiously, primarily serving as a pragmatic way to reconcile explanation additivity with the target prediction when fidelity is imperfect \citep{Rudin2019Stop, Yeh2019Faithfulness}. Hence it is optional depends on the XAI system requirements.

\newpage

\bibliographystyle{unsrtnat}
\bibliography{references}

\end{document}